\documentclass{article}
\usepackage{spconf,amsmath,graphicx}
\usepackage{amssymb}
\usepackage{multirow}
\usepackage{colortbl}
\usepackage{xcolor}
\usepackage{booktabs}
\usepackage{subcaption}
\usepackage{enumitem}
\usepackage{hyperref}   
\usepackage{cleveref}

\title{Edit2Interp: Adapting Image Foundation Models from Spatial Editing to Video Frame Interpolation with Few-Shot Learning}
\name{\begin{tabular}{c}
Nasrin Rahimi$^{\star}$, M{\i}sra Yavuz$^{\star}$, Burak Can Biner$^{\star}$, Yunus Bilge Kurt$^{\star}$, 
Ahmet Rasim Emirda\u{g}{\i}$^{\star}$, \\ S\"{u}leyman Aslan$^{\star}$, G\"{o}rkay Aydemir$^{\star}$, M. Ak{\i}n Y{\i}lmaz$^{\star}$, A. Murat Tekalp$^{\dagger}$
\end{tabular}
\thanks{This work is supported by Codeway.}}
\address{$^{\star}$ Codeway AI Research \\
$^{\dagger}$ Dept. of Electrical \& Electronics Engineering, Ko\c{c} University, \.{I}stanbul, T\"{u}rkiye}
\begin{document}

%\ninept
%
\maketitle
\begin{abstract}
Pre-trained image editing models exhibit strong spatial reasoning and object-aware transformation capabilities acquired from billions of image-text pairs, yet they possess no explicit temporal modeling. This paper demonstrates that these spatial priors can be repurposed to unlock temporal synthesis capabilities through minimal adaptation—without introducing any video-specific architecture or motion estimation modules. We show that a large image editing model (Qwen-Image-Edit), originally designed solely for static instruction-based edits, can be adapted for Video Frame Interpolation (VFI) using only 64-256 training samples via Low-Rank Adaptation (LoRA). Our core contribution is revealing that the model's inherent understanding of "how objects transform" in static scenes contains latent temporal reasoning that can be activated through few-shot fine-tuning. While the baseline model completely fails at producing coherent intermediate frames, our parameter-efficient adaptation successfully unlocks its interpolation capability. Rather than competing with task-specific VFI methods trained from scratch on massive datasets, our work establishes that foundation image editing models possess untapped potential for temporal tasks, offering a data-efficient pathway for video synthesis in resource-constrained scenarios. This bridges the gap between image manipulation and video understanding, suggesting that spatial and temporal reasoning may be more intertwined in foundation models than previously recognized.

\end{abstract}
\section{Introduction}
\label{sec:intro}
\vspace{-3pt}

Video Frame Interpolation (VFI) is a pivotal task in computer vision~\cite{park2021ABME,l2bec2,ttvfi,gimmvfi}, aimed at synthesizing coherent intermediate frames between consecutive video frames. It is an essential tool for high-frame-rate video conversion, slow-motion synthesis, and temporal super-resolution. Traditionally, the field has been dominated by architectures that rely on explicit motion estimation, such as optical flow-based warping or spatially-adaptive convolution kernels~\cite{enbmvfi, edcvfi,emahqvfi}. While these methods have achieved significant milestones, they often struggle with complex non-linear motions, occlusions, and the synthesis of realistic textures in challenging scenarios~\cite{eden,sgmvfi, ocai}. Furthermore, these specialized models typically require training on massive, high-bitrate datasets like Vimeo-90K~\cite{vimeo90k} to learn motion priors from scratch. Despite substantial progress, the reliance on explicit motion estimation and large-scale task-specific supervision fundamentally constrains the expressiveness and data efficiency of existing VFI methods, motivating the exploration of more general-purpose visual models with richer semantic priors.

In this context, general-purpose visual foundation models demonstrate strong generalization across downstream tasks in zero- and few-shot settings. In particular, the emergence of large-scale visual foundation models (VFMs), particularly generative image editing models such as Flux-kontext~\cite{labs2025flux} and Qwen-Image-Edit~\cite{qwen}, has introduced a new paradigm in visual synthesis. These models are pre-trained on billions of image-text pairs, granting unparalleled spatial reasoning and semantic ``object-aware'' priors. However, such models are fundamentally designed for static image manipulation via textual instructions, and they lack the temporal consistency required for video-level tasks. While their baseline fails to produce coherent intermediate frames out-of-the-box, their rich internal representations of ``how objects change'' suggest a latent potential for temporal tasks. 

\begin{figure}[t!]
    \centering
    \includegraphics[width=\columnwidth]{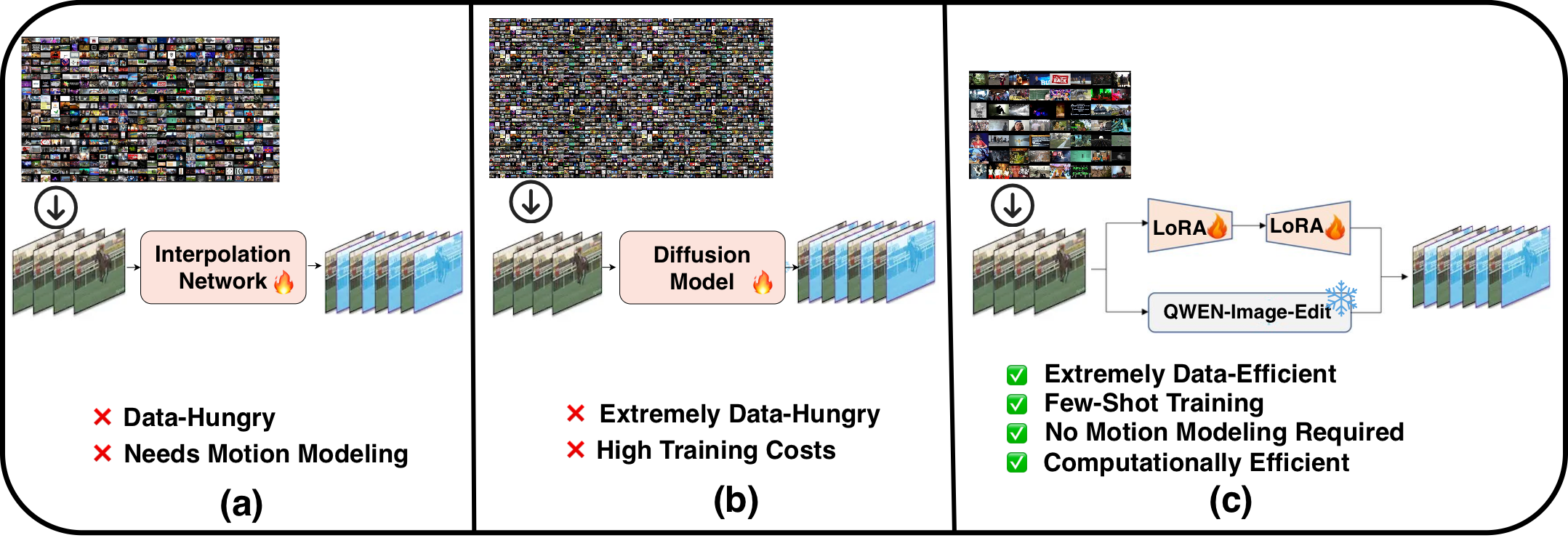}
    \captionsetup{font=small}
    \caption{Conceptual comparison of VFI paradigms. (a) Traditional VFI networks,(b) Task-specific diffusion models, and (c) Our Edit2Interp approach}
    \label{fig:comparison}
    \vspace{-6pt}
\end{figure}

In this work, we propose a novel framework to bridge the gap between static image editing and temporal interpolation. 
We reframe VFI as a constrained image editing task, where the goal is to synthesize the intermediate frame conditioned on both the first and last frames, along with a textual prompt. To maintain efficiency and leverage the power of foundation models, we utilize Low-Rank Adaptation (LoRA)~\cite{lora} to adapt the Qwen-Image-Edit model to this task in an extreme few-shot regime. By using as few as 64 training samples, we ``unlock'' the model's ability to perform temporal inbetweening while preserving the high-fidelity generation capabilities inherent in its diffusion-based architecture. 
We contrast our suggested paradigm with existing approaches in \textbf{Fig. \ref{fig:comparison}}. While traditional VFI architectures often rely on explicit motion estimation and are data-hungry (Fig. \ref{fig:comparison}a), recent task-specific diffusion models, though perceptually powerful, remain extremely data-hungry with high training costs (Fig. \ref{fig:comparison}b). In contrast, our \textbf{Edit2Interp} approach (Fig. \ref{fig:comparison}c) highlights a shift toward extreme data efficiency and computational simplicity by repurposing pre-trained spatial priors through lightweight LoRA adaptation and prioritizes generative realism and structural consistency.

We summarize our main contributions below: 
\begin{itemize}[nosep,leftmargin=14pt,labelindent=6pt]
    \item We introduce a method to repurpose large-scale image editing foundation models for VFI, treating temporal synthesis as a constrained editing task. 
    \item We demonstrate that high-fidelity VFI can be achieved through few-shot adaptation, requiring as low as 64 samples—a significant reduction compared to the tens of thousands of samples required by traditional SOTA models. 
    \item We show that Parameter-Efficient Fine-Tuning (PEFT) via LoRA can successfully ``unlock'' the temporal logic in a model natively trained only for static image edits and camera-angle controls. 
    \item Our approach leverages the strong spatial priors of the foundation models to synthesize temporally plausible intermediate frames, yielding substantial improvements over the base model and competitive perceptual quality under a few-shot adaptation regime, despite not being explicitly trained for video understanding.
\end{itemize}

\vspace{-3pt}
\section{Related Works}
\label{sec:related}
\vspace{-3pt}

\noindent \textbf{Flow-based and Kernel-based Methods.}
Flow-based approaches typically involve bidirectional flow estimation followed by warping. Recent methods enhance interpolation by utilizing occlusion and consistency-aware mechanisms~\cite{ocai}. To alleviate the rigid assumptions of traditional motion modeling, ~\cite{gimmvfi} employs coordinate-based neural networks for arbitrary-timestep flow prediction, while~\cite{enbmvfi} refines the pyramid recurrent framework across varied resolutions. On the other hand, kernel-based methods synthesize pixels by applying spatially adaptive kernels.~\cite{emahqvfi} addresses motion ambiguity and alignment using a bidirectional motion field. \cite{edcvfi} utilizes a coarse-to-fine 3D CNN to improve multi-flow prediction. \cite{l2bec2} introduces a local lightweight bidirectional encoding with a channel attention cascade for improved efficiency.

\noindent \textbf{Transformer-based Methods.}
Transformers have become the state-of-the-art for capturing long-range dependencies.~\cite{sgmvfi} addresses large motion challenges by integrating global level information, while~\cite{emavfi} proposes a unified module to disentangle motion and appearance features and \cite{ttvfi} introduces a trajectory aware transformer to mitigate artifacts in complex non-linear motion.

\noindent \textbf{Diffusion-based Methods.}
The emergence of generative models has pushed the boundaries of perceptual realism in VFI. \cite{vdim}~demonstrates the~power of diffusion models in handling motion ambiguity. \cite{danier2023ldmvfi}~performs denoising in the latent space to produce high-fidelity results. To improve sampling efficiency and quality,~\cite{vibidsampler} uses a bidirectional diffusion sampler, while~\cite{hierdif} explicitly models optical flow to reduce the search space. To tackle extreme motion,~\cite{eden} proposes an enhanced diffusion framework designed for high-quality large-motion interpolation.

\noindent \textbf{Few-shot Adaptation and Foundation Models.}
Traditional VFI architectures are typically trained from scratch on tens of thousands of video clips to learn motion dynamics. In contrast, we explore Parameter-Efficient Fine-Tuning (PEFT) of Visual Foundation Models like Qwen Image Edit, which possess extensive spatial and semantic knowledge. 
Several PEFT strategies have been proposed~\cite{lora,waveft}. LoRA~\cite{lora} introduces trainable low-rank matrices into transformer layers, significantly reducing the number of parameters updated. Building upon this,~\cite{shira} utilizes sparse high rank adaptation process, while~\cite{waveft} leverages wavelet-based fine-tuning to better capture multi-frequency spatial details during adaptation. Similarly,\cite{yilmaz2026edit2restore} adapts image foundation models for few-shot image restoration tasks.

Our approach is distinct in its extreme data efficiency. While most generative VFI models require large-scale supervision, we demonstrate that by applying LoRA to Qwen Image Edit, we can achieve robust interpolation in a few-shot design. This enables the model to repurpose its image-editing logic into a temporal interpolator, a capability that is entirely absent in the original baseline foundation model.

\vspace{-6pt}
\section{Method}
\label{sec:method}
Following the paradigm shift established in Section 1 (Fig. \ref{fig:comparison}), our framework departs from motion-centric interpolation pipelines by repurposing a frozen image foundation model for VFI via lightweight adaptation. In this section, we first introduce the Qwen-Image-Edit backbone, then describe our constrained editing formulation for VFI, and present the few-shot LoRA adaptation strategy.

\vspace{-6pt}
\subsection{Foundation Model Backbone: Qwen-Image-Edit}
\label{ssec:backbone}
\vspace{-3pt}

Our framework leverages the \textbf{Qwen-Image-Edit-2509}~\cite{qwen} architecture, a Large-scale Multi-Modal Diffusion Transformer (MMDiT) denoted as a function $\mathcal{F}_\theta$ with pre-trained weights $\theta$.  
This backbone is designed for instruction-based image manipulation using $N+1$ conditioning images $\{I^{(n)}\}_{n=0}^N$ and an instruction prompt $\mathcal{P}$. We denote the latent representations of the conditioning images by $\{z^{(n)}\}_{n=0}^N$, where $z^{(n)} = \mathcal{E}_{\mathrm{VAE}}(I^{(n)})$, and $\mathcal{E}_{\mathrm{VAE}}$ denotes the VAE encoder.

To perform synthesis, the model $\mathcal{F}_\theta$ predicts the velocity field $v_t$ to denoise the latent $z_t$ at diffusion timestep $t \in [0,1]$, conditioned on $\mathbf{C}$, i.e., $v_t = \mathcal{F}_\theta(z_t, t, \mathbf{C})$.

The backbone model utilizes a dual-conditioning mechanism. To obtain reconstructive features, it uses the latent representatio of the conditioning images, $\{z^{(n)}\}_{n=0}^N$, In parallel, a frozen Qwen2.5-VL encoder denoted by $\phi$ extracts semantic tokens $h$ from the conditioning images and a constant textual instruction $\mathcal{P}$, i.e., $h = \phi(\mathcal{P}, \{I^{(n)}\}_{n=0}^N)$. The final conditioning signal is therefore formulated as:
\begin{equation}
    \mathbf{C} = \{h,\{z^{(n)}\}_{n=0}^N\}
\end{equation}

Through cross-attention and multimodal modulation layers, the model fuses these modalities to form an ``object-aware'' representation of scene geometry. Although natively static, its rich internal representations provide the structural priors needed for complex motion synthesis.

\begin{figure*}[t!]
    \centering
    \includegraphics[width=0.8\textwidth]{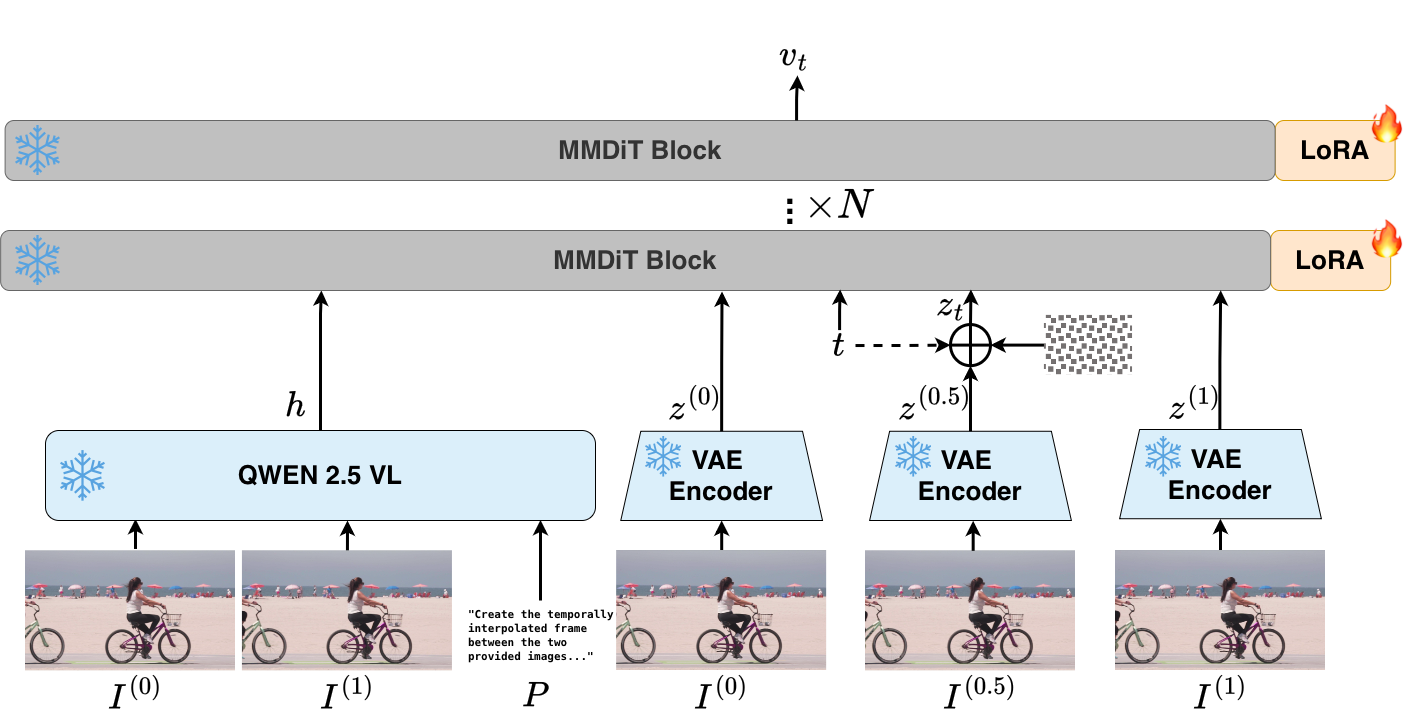}
    \captionsetup{font=small}
    \caption{Overview of the Edit2Interp framework}
    \label{fig:pipeline}
    \vspace{-6pt}
\end{figure*}

\vspace{-6pt}
\subsection{Repurposing for VFI via Constrained Editing and LoRA}
\label{ssec:reformulation_lora}
\vspace{-3pt}

We repurpose the aforementioned image-editing backbone for VFI. 
Given two boundary frames $\{I^{(0)}, I^{(1)}\}$, the objective of VFI is to synthesize the intermediate frame $I^{(0.5)}$. 
We reformulate this task as a constrained multimodal editing problem.
As illustrated in Figure \ref{fig:pipeline}, in our proposed \textbf{Edit2Interp} pipeline, the boundary frames 
$\{I^{(0)}, I^{(1)}\}$ together with a fixed textual instruction $\mathcal{P}$ constitute the conditioning inputs.  $I^{(0)}$ provides the initial spatial ``canvas'', while $I^{(1)}$ acts as a visual boundary condition that specifies the temporal destination, and, $\mathcal{P}$ triggers the model's transformation logic.
Following the conditioning mechanism defined in Section~\ref{ssec:backbone}, the spatial latent conditions are obtained by encoding the boundary frames into the latent space, yielding $z^{(0)}$ and $z^{(1)}$. The semantic tokens $h$ are computed using Qwen2.5-VL encoder $\phi$:
\vspace{-6pt}
\begin{equation}
    h = \phi(\mathcal{P}, \{I^{(0)}, I^{(1)}\}).
\end{equation}
The resulting conditioning set is
\vspace{-6pt}

\begin{equation}
    \mathbf{C} = \{h, z^{(0)}, z^{(1)}\}.
\end{equation}
The synthesis of $I_{0.5}$ is thus modeled as an in-betweening edit, where the model must resolve the logical midpoint $z_{0.5}$ by reconciling the spatial latents and semantic tokens of the two boundary frames. As shown in Fig.~\ref{fig:pipeline}, during adaptation phase, $z_{0.5}$ serves as the target, where we supervise the model's ability to recover this midpoint latent after it has been perturbed with Gaussian noise.
This approach bypasses the need for explicit optical flow, relying instead on the model's latent ability to perform morphological transformation.

To activate the model’s latent interpolation capability in a parameter-efficient manner while preserving generative fidelity, we employ LoRA~\cite{lora}. We freeze the VAE, Qwen2.5-VL encoder, and all original weights $\theta$, optimizing only the LoRA parameters $\Delta\theta$. For a frozen weight matrix $W_0$, LoRA introduces trainable matrices $A$ and $B$, yielding the modified output $h = W_0 x + \frac{\alpha}{r} B A x$, where $x$ is the layer input, $\alpha$ is a scaling factor, and $r$ is the LoRA rank. We apply LoRA to MMDiT blocks to ensure deep multimodal alignment.

The adaptation follows a Rectified Flow Matching~\cite{esser2024scaling} objective, where the target middle-frame latent $z^{(0.5)}$ is perturbed with Gaussian noise:
\vspace{-6pt}
\begin{equation}
    z_t = (1 - t)\, z^{(0.5)} + t\, \epsilon,
    \quad \epsilon \sim \mathcal{N}(0, I),
    \quad t \in [0,1]
\end{equation}

The model learns to predict the velocity field $v_t = \epsilon - z^{(0.5)}$ required to transform the noisy latent back to the target. The LoRA-enhanced model is optimized via:
\vspace{-6pt}
\begin{equation}
    \mathcal{L}=\mathbb{E}_{(z^{(0.5)}, \mathbf{C}) \sim \mathcal{D},\, \epsilon,\, t}[||\mathcal{F}_{\theta + \Delta\theta}(z_{t},t,\mathbf{C})-v_{t}||^{2}]
\end{equation}
where $\mathcal{F}_{\theta + \Delta\theta}$ the LoRA-enhanced model and $\mathcal{D}$ denotes the training dataset.
In the inference, synthesis begins from pure Gaussian noise, where the model iteratively integrates the learned velocity field to reconstruct $z_{0.5}$.
By fine-tuning on extremely small data regimes ($\{64, 128, 256\}$), we investigate whether the model can ``unlock'' temporal reasoning by simply realigning its pre-existing spatial knowledge to the interpolation task.

\vspace{-6pt}
\section{Experiments and Results}
\vspace{-3pt}
\subsection{Experimental Setup}
\noindent \textbf{Datasets.}
For few-shot adaptation, we randomly sample 64, 128, and 256 triplets from the Vimeo-90K~\cite{vimeo90k} training set. To evaluate generalization, we additionally employ a multi-dataset training regime incorporating high-resolution triplets from DAVIS~\cite{davis}, Middlebury~\cite{baker2011database}, and UVG~\cite{uvg}. 
We evaluate the baseline and adapted variants on three benchmarks: Vimeo-90K (test), UCF101~\cite{soomro2012ucf101}, and DAVIS (test), which span diverse motion patterns, resolutions, and aspect ratios for comprehensive generalization assessment.

\noindent \textbf{Implementation Details.}
We implement our framework in PyTorch and run all experiments on a single NVIDIA H100 GPU. The pre-trained Qwen-Image-Edit-2509 is adapted by training LoRA adapters on the attention projections ($q,k,v$), output layers, and modulation MLPs across MMDiT blocks. We use AdamW~\cite{adamw} with a constant learning rate of $1\times10^{-4}$ for 10 epochs in \texttt{bfloat16} precision. 
During inference, we use the same fixed text prompt for consistent conditioning and employ a Flow Matching scheduler with discrete Euler steps, 40 denoising steps, and a classifier-free guidance scale of 1.

\noindent \textbf{Evaluation Metrics.}
We evaluate synthesized intermediate frames using complementary metrics capturing pixel accuracy, perceptual realism, and temporal consistency, with emphasis on perceptual and temporal measures of visual realism and motion coherence. In addition to PSNR, we report Fréchet Inception Distance (FID)~\cite{fid} for distributional similarity between generated and ground-truth frames, and Learned Perceptual Image Patch Similarity (LPIPS)~\cite{lpips} for feature-level consistency. 
To further assess temporal coherence, we report FloLPIPS~\cite{danier2022flolpips} and Perceptual Straightness (PS)~\cite{rahimi2023spatio}, which capture temporal artifacts and motion inconsistencies not reflected by frame-wise metrics.

\vspace{-3pt}
\subsection{Quantitative Results}
\vspace{-3pt}
%\textbf{The Unlock Effect: Zero-shot vs. Few-shot Performance}
Table~\ref{tab:quant_results} reports the performance of the frozen Qwen-Image-Edit backbone (Baseline) and our LoRA-adapted variants trained with different ranks and data regimes on the Vimeo-90K, UCF101, and DAVIS benchmarks. For contextual comparison, we include the task-specific diffusion model LDMVFI~\cite{danier2023ldmvfi}, trained from scratch on large-scale VFI data, as a reference point.
As shown in Table~\ref{tab:quant_results}, the Baseline performs poorly across all datasets, highlighting the domain gap between static image editing and VFI tasks. In contrast, LoRA fine-tuning yields substantial improvements over the frozen model across all metrics, even with as few as 64 video triplets. These results show that limited task-specific temporal supervision is sufficient to activate the latent interpolation capability of the pre-trained image foundation model.
While the task-specific LDMVFI model, trained from scratch for VFI, achieves superior results on all metrics, our method produces perceptually coherent and temporally smooth interpolations that are  comparable, using minimal supervised fine-tuning of a pre-trained image model.

\begin{table}[!t]
\centering
\captionsetup{font=small}
\caption{Quantitative comparison on different datasets.best in bold, second best underlined.}  \vspace{-6pt}
\label{tab:quant_results}
\setlength{\tabcolsep}{3.5pt}
\small
\resizebox{\columnwidth}{!}{%
\begin{tabular}{llccccc}
\toprule
\textbf{Dataset} & \textbf{Method} & \textbf{PSNR $\uparrow$} & \textbf{LPIPS $\downarrow$} & \textbf{FID $\downarrow$} & \textbf{FloLPIPS $\downarrow$} & \textbf{PS $\uparrow$}\\
\midrule

\multicolumn{6}{l}{\textbf{VIMEO90k}} \\
\midrule
 & Baseline (image edit) & 19.80 & 0.207 & 11.26 & 0.258 & 51.53 \\
 & LoRA-r64-N64    & 23.63 & 0.074 & 4.38 & 0.104 & 88.29 \\
 & LoRA-r64-N128   & 26.17 & 0.053 & 3.50 & 0.766 & 100.35 \\
 & LoRA-r128-N64   & 24.01 & 0.066 & 3.94 & 0.095 & 90.59 \\
 & LoRA-r128-N128  & 26.66 & 0.051 & 3.47 & 0.072 & 101.04 \\
 & LoRA-r128-N256  & \underline{27.60} & \underline{0.047} & \underline{3.25} & \underline{0.066} & \underline{105.96} \\
 & LoRA-r128-N256-Multi & \textbf{27.83} & \textbf{0.041} & \textbf{2.73} & \textbf{0.058} & \textbf{108.82} \\
 
 \rowcolor{gray!15}
 & LDMVFI~\cite{danier2023ldmvfi}(VFI task-specific) & 33.09 & 0.023 & 2.49 & 0.045 & 121.93 \\

\midrule
\multicolumn{6}{l}{\textbf{UCF101}} \\
\midrule
 & Baseline (image edit) & 19.08 & 0.295 & 74.98 & 0.318 & 33.77 \\
 & LoRA-r64-N64    & 30.02 & 0.043 & 20.27 & 0.059 & 78.31 \\
 & LoRA-r64-N128   & 30.99 & 0.036 & 17.29 & 0.048 & 83.39 \\
 & LoRA-r128-N64   & 30.52 & 0.039 & 16.94 & 0.052 & 80.22 \\
 & LoRA-r128-N128  & 30.96 & 0.037 & 16.95 & 0.048 & 84.86 \\
 & LoRA-r128-N256  & \textbf{31.73} & \textbf{0.034} & \textbf{15.48} & \textbf{0.044} & \underline{90.27} \\
 & LoRA-r128-N256-Multi & \underline{31.71} & \underline{0.034} & \underline{16.36} & \underline{0.045} & \textbf{95.65} \\
 
 \rowcolor{gray!15}
 & LDMVFI~\cite{danier2023ldmvfi}(VFI task-specific) & 34.70 & 0.024 & 12.93 & 0.032 & 126.43 \\

\midrule
\multicolumn{6}{l}{\textbf{DAVIS}} \\
\midrule
 & Baseline (image edit) & 18.10 & 0.275 & 22.67 & 0.323 & 58.72 \\
 & LoRA-r64-N64    & 19.92 & 0.201 & 10.45 & 0.238 & 82.95 \\
 & LoRA-r64-N128   & 21.10 & 0.180 & 10.45 & 0.215 & 90.98 \\
 & LoRA-r128-N64   & 19.84 & 0.192 & \underline{10.09} & 0.226 & 83.06 \\
 & LoRA-r128-N128  & 21.38 & 0.177 & 10.63 & 0.211 & 92.38 \\
 & LoRA-r128-N256  & \underline{22.04} & \underline{0.170} & 10.52 & \underline{0.203} & \underline{97.04} \\
 & LoRA-r128-N256-Multi & \textbf{22.26} & \textbf{0.135} & \textbf{8.03} & \textbf{0.164} & \textbf{98.50} \\

 \rowcolor{gray!15}
 & LDMVFI~\cite{danier2023ldmvfi}(VFI task-specific) & 25.32 & 0.115 & 14.59 & 0.159 & 111.39 \\
\bottomrule
\end{tabular}
}
\vspace{-12pt}
\end{table}

\vspace{-3pt}
\subsection{Visual Results}
\vspace{-3pt}

\begin{figure}[tb]
\centering
\setlength{\tabcolsep}{1pt}

\vspace{2pt}
\footnotesize (a) Vimeo90K \\
\includegraphics[width=\columnwidth]{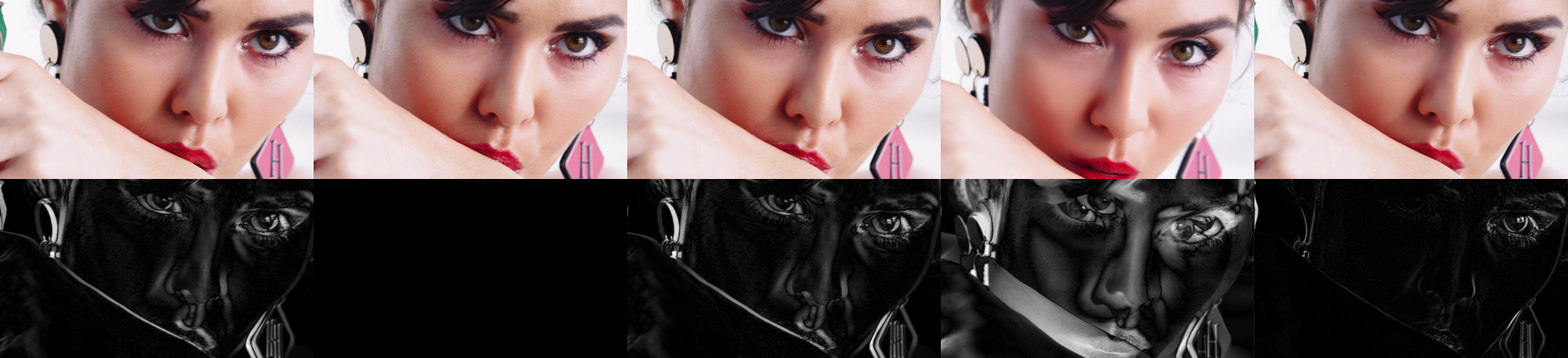} \\

\vspace{4pt}
\footnotesize (b) UCF101 \\
\includegraphics[width=\columnwidth]{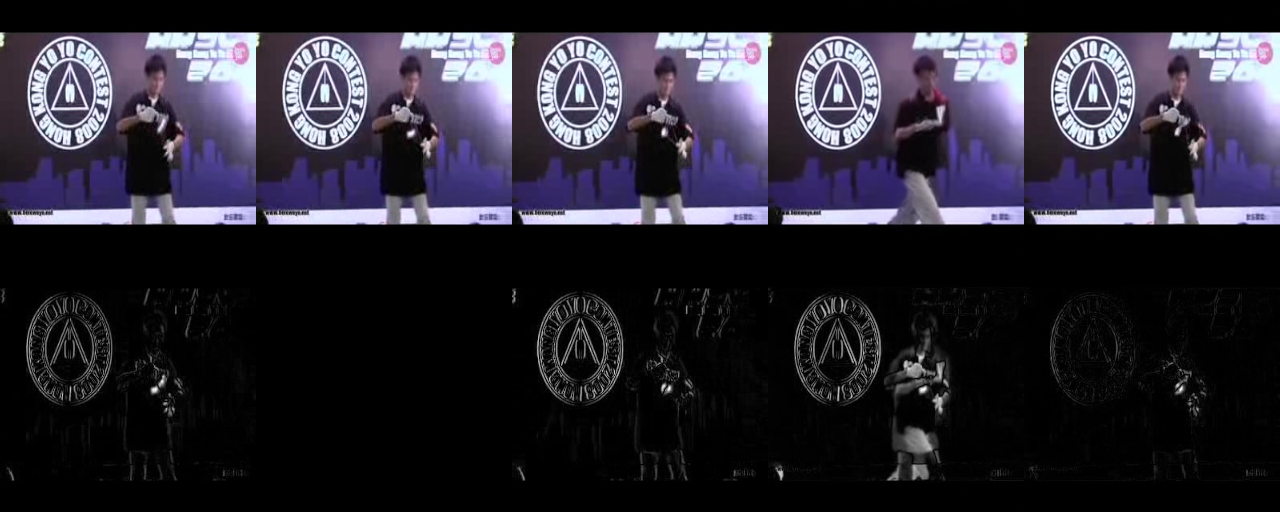} \\

\vspace{4pt}
\footnotesize (c) DAVIS 
\includegraphics[width=\columnwidth]{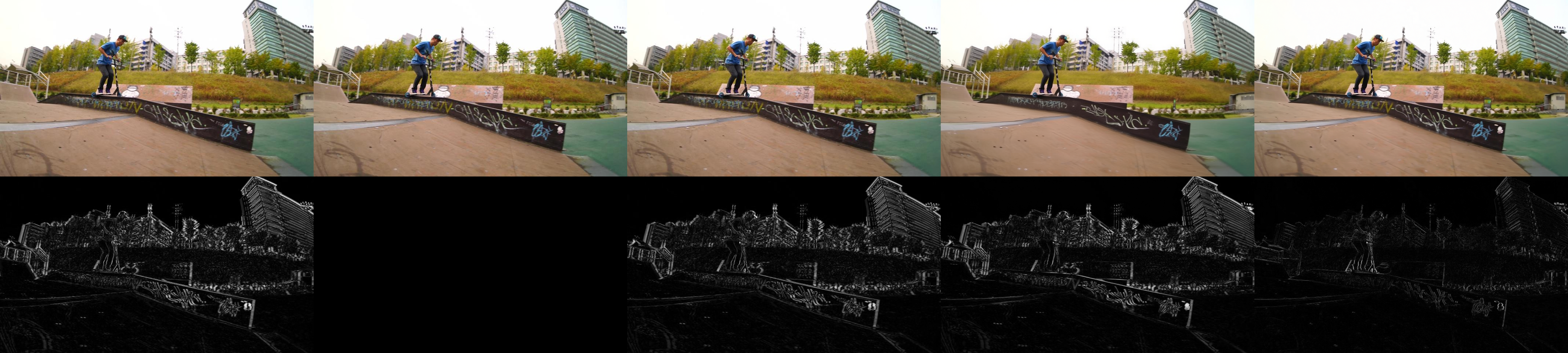} \\
    \vspace{-7pt}
\captionsetup{font=small}
\caption{
Visual comparison across datasets.
Top row for each dataset: ground truth frames $t\!-\!1$, $t$, $t\!+\!1$, baseline interpolation $t$, and LoRA interpolation $t$.
Bottom row for each dataset: absolute difference maps with respect to the ground truth frame $t$.
}
\label{fig:qualitative}
\vspace{-6pt}
\end{figure}

Figure~\ref{fig:qualitative} presents a qualitative comparison between \textbf{LoRA-r128-N256} and the frozen baseline model for interpolating the middle frame $t$ across the Vimeo-90K, UCF101, and DAVIS datasets. 
Across all datasets, the LoRA-adapted model generates intermediate frames that are visually closer to the ground truth, demonstrating improved structural alignment and appearance fidelity. This is reflected in the reduced magnitude and spatial extent of the absolute difference maps with respect to the ground truth frame $t$. In contrast, the baseline exhibits larger residual errors and visible artifacts, indicating its limited ability to model temporal transformations.
Displaying absolute difference maps of the boundary ground-truth frames ($t!-!1$, $t!+!1$) confirms that the LoRA model does not collapse to either boundary frame, as evidenced by the reduced error magnitudes.

A key qualitative distinction lies in motion consistency. The frozen baseline lacks an explicit temporal understanding and often produces misaligned structures or unstable transitions. By contrast, the LoRA-tuned model better preserves object geometry and motion continuity, resulting in smoother and more coherent interpolations.
\vspace{-6pt}

\vspace{-3pt}
\subsection{Ablation Studies}
\vspace{-3pt}

We analyze the effect of LoRA capacity and training data size on interpolation performance by varying the LoRA rank $r \in \{64, 128\}$ and the number of training triplets $N \in \{64, 128, 256\}$, with the results summarized in Table~\ref{tab:quant_results}.

\noindent \textbf{Effect of LoRA Rank.}
For a fixed number of training samples, increasing the LoRA rank consistently improves performance across datasets.
For example, on Vimeo-90K with $N=128$, increasing the rank from $r=64$ to $r=128$ improves PSNR from 26.17~dB to 26.66~dB while simultaneously reducing LPIPS and FID.
This trend indicates that higher-rank adapters provide additional expressive capacity to model complex temporal correspondences, while still preserving the strong spatial priors of the frozen backbone.

\noindent\textbf{Effect of Training Data Size.}
For a fixed LoRA rank, increasing the number of training triplets leads to monotonic improvements in both distortion-based and perceptual metrics.
Notably, even with only $N=64$ training samples, LoRA adaptation yields a significant improvement over the frozen baseline, particularly in perceptual metrics such as LPIPS and FID, demonstrating strong data efficiency.
Performance continues to improve as $N$ increases to 256, suggesting that the model benefits from additional temporal supervision without overfitting in this low-data regime.

\noindent \textbf{Cross-Dataset Consistency.}
The observed trends are consistent across Vimeo-90K, UCF101, and DAVIS, despite differences in resolution, aspect ratio, and motion characteristics.
Models adapted on small subsets of Vimeo-90K generalize well to UCF101 and DAVIS, indicating that LoRA primarily aligns the foundation model’s pre-trained visual representations with the temporal interpolation task rather than learning dataset-specific motion patterns.

\noindent \textbf{Multi-Dataset Adaptation}
While models adapted on small subsets of Vimeo-90K show cross-dataset consistency, we further evaluated generalization by training a multi-dataset variant, denoted as LORA-r128-N256-Multi, using 256 triplets sampled from Vimeo-90K, DAVIS, Middlebury, and UVG. Results show that training on high-resolution data (e.g., UVG and DAVIS) significantly improves generalization to similar high-quality test sets. However, we identified an 
``artifact barrier'': generalization to lower-quality datasets like UCF101 is limited when training data is ``too clean''. The model prioritizes the high-fidelity reconstruction logic of the foundation backbone, making it less robust to the compression artifacts found in low-bitrate or older video samples.

\vspace{-6pt}
\section{Conclusion}
\label{sec:conclusion}
\vspace{-6pt}
In this paper, we introduce \textbf{Edit2Interp}, a framework that repurposes foundation image editing diffusion models for VFI through minimal adaptation. By leveraging LoRA, we show that a foundation model natively designed for static instruction-based image editing can be effectively aligned with temporal reasoning, achieving substantial performance gains even in extreme few-shot settings with as few as 64--256 training triplets.
Our results indicate that such foundation diffusion models already encode rich spatial and semantic priors, which can be redirected toward interpolation without explicit motion estimation or large-scale video supervision. While our approach does not match flow-based methods in quantitative metrics, it achieves competitive visual perceptual quality and generalizes well across datasets, reflecting a fundamental trade-off between motion-based modeling and generative interpolation via image foundation models. A current limitation of our method is inference latency, which can be overcome by accelerated diffusion sampling methods. 

This paper demonstrates that minimally adapted image foundation models offer a flexible and highly data-efficient alternative to task-specific video interpolation architectures. This work represents a step forward in the direction of unifying image and video foundation models, blurring the boundaries between image manipulation and video synthesis, and opening affordable fine-tuning options for high-quality video enhancement in data-constrained environments.

\clearpage
\let\oldbibliography\bibliography
\renewcommand{\bibliography}[1]{%
\begin{small} % Reduces font size to 'small'
  \oldbibliography{#1}
\end{small}
}
\bibliographystyle{IEEEbib}
\bibliography{refs}

\end{document}